\documentclass[conference]{IEEEtran}
\ifCLASSINFOpdf
\else
\fi

\usepackage{graphicx} 
\usepackage{amsmath} 
\usepackage{amssymb}  
\usepackage[hidelinks, bookmarks=false]{hyperref}
\usepackage{subcaption}
\usepackage{comment}
\usepackage[english, ruled, linesnumbered, vlined]{algorithm2e}
\usepackage{booktabs}
\usepackage{bm}

\begin{document}
%
\title{This Far, No Further: Introducing Virtual Borders to Mobile Robots Using a Laser Pointer}

\author{\IEEEauthorblockN{Dennis Sprute}
\IEEEauthorblockA{Bielefeld University of Appl. Sciences \\
Campus Minden\\
32427 Minden, Germany\\
Email: dennis.sprute@fh-bielefeld.de}
\and
\IEEEauthorblockN{Klaus T\"{o}nnies}
\IEEEauthorblockA{Otto-von-Guericke University Magdeburg \\
Faculty of Computer Science\\
39106 Magdeburg, Germany\\
Email: klaus@isg.cs.uni-magdeburg.de}
\and
\IEEEauthorblockN{Matthias K\"{o}nig}
\IEEEauthorblockA{Bielefeld University of Appl. Sciences \\
Campus Minden\\
32427 Minden, Germany \\
Email: matthias.koenig@fh-bielefeld.de}}


%


\maketitle

\begin{abstract}
We address the problem of controlling the workspace of a 3-DoF mobile robot. In a human-robot shared space, robots should navigate in a human-acceptable way according to the users' demands. For this purpose, we employ virtual borders, that are non-physical borders, to allow a user the restriction of the robot's workspace. To this end, we propose an interaction method based on a laser pointer to intuitively define virtual borders. This interaction method uses a previously developed framework based on robot guidance to change the robot's navigational behavior. Furthermore, we extend this framework to increase the flexibility by considering different types of virtual borders, i.e. polygons and curves separating an area. We evaluated our method with 15 non-expert users concerning correctness, accuracy and teaching time. The experimental results revealed a high accuracy and linear teaching time with respect to the border length while correctly incorporating the borders into the robot's navigational map. Finally, our user study showed that non-expert users can employ our interaction method.
\end{abstract}


%
\IEEEpeerreviewmaketitle

\section{Introduction}
Robots pervasively find their ways into human-centered environments, such as home environments, and support the people in their everyday life. They act as service robots vacuuming the house or as companion robots supporting the residents. This help is greatly appreciated by humans. However, from our experience we know that there are situations in which we want to restrict the workspace of a mobile robot. An everyday scenario is the restriction of the workspace of a vacuum cleaning robot. The user does not want the robot to clean the carpet on the ground and wants the robot to circumvent the carpet while vacuuming the rest of the room. Furthermore, residents want to prevent the robot of entering certain social places, e.g. a bath room or an area of a room due to privacy concerns. Thus, mobile robots should navigate in a human-acceptable way respecting the user-defined workspace.\par

For this purpose, we employ \textit{virtual borders} that are respected by the mobile robot. In contrast to physical borders, e.g. walls or furniture, virtual borders are not directly visible to the user but indicate occupied areas to the robot. The teaching process of such virtual borders should be flexible and applicable by non-expert users. Such a non-expert (1)~neither has any programming skills (2)~nor experience with or insights into robotics, but (3)~gets along with common consumer devices, e.g. smartphones, tablets and presenters. Furthermore, a non-expert user (4)~does not care about details of a system but is rather interested in a robust and feature-complete system. Finally, we assume a non-expert to (5)~have no cognitive impairments or upper limb disorders.\par
\begin{figure}
	\centering
	\includegraphics[width=0.45\textwidth]{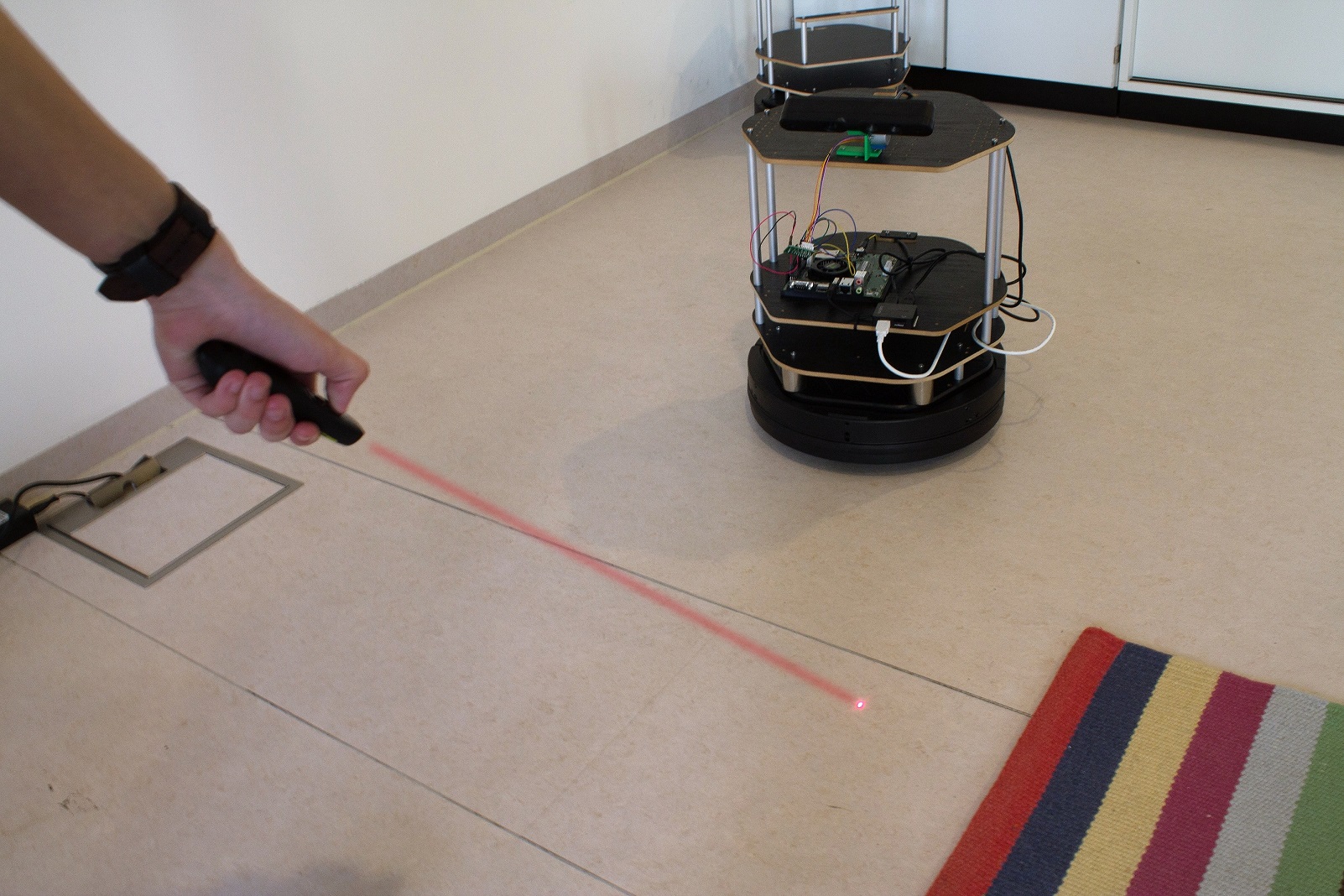}   
	\caption{A user restricts the robot's workspace by showing virtual borders with a laser pointer.}
	\label{fig:motivation}
\end{figure}

In order to address such a non-expert user and system requirements concerning high accuracy, little teaching effort, low intrusiveness and high flexibility, different methods are imaginable (see Sect.~\ref{sec:relatedWork}). We build on a previously developed framework that allows the definition of virtual borders based on trajectories of a mobile robot guided by a human teacher~\cite{Sprute:2017}. Several human-robot interfaces could be used to guide the robot and define the virtual borders, e.g. direct physical control, remote controllers, smartphones, tablets, pointing gestures (with or without auxiliary device). In this paper, we contribute \textbf{(1) an interaction method based on a laser pointer} as human-robot interaction device for teaching virtual borders. A laser pointer is more accurate in border teaching compared to human gestures due to the inherent uncertainty in accurate gesture recognition. Rouanet et al. also showed that especially non-experts judge mediator devices to be more intuitive and efficient in teaching new visual objects compared to human gestures~\cite{Rouanet:2013}. Besides, we think guiding a robot with a laser pointer is easier for non-experts compared to remote controllers, smartphones or tablets. Users need to put themselves into the position of the robot when using one of the latter devices, whereas using a laser pointer only requires providing a position on the ground plane without considering the perspective of the robot. Moreover, the laser spot on the ground provides inherently visual feedback to the user. Finally, since laser pointers are common in our everyday live, e.g. oral presentations, they are intuitive interaction devices. Fig.~\ref{fig:motivation} illustrates a scenario where a user guides a robot around a carpet to exclude the carpet area from its workspace. Moreover, our interaction method comprises a simple, yet powerful, feedback system, that only relies on the mobile robot and no additional devices. Additional to the proposed interaction method, we contribute \textbf{(2) an extension to the employed teaching framework} to enhance its flexibility by distinguishing between different border types, i.e. polygons and curves separating an area.

\section{Related Work}
\label{sec:relatedWork}
Occupancy grid maps model the environment by means of cells containing a probability for the occupancy of the corresponding area~\cite{Moravec:1985}. An overview of different map representations is given by Fuentes-Pacheco et al.~\cite{Fuentes-Pacheco:2015}. SLAM algorithms \cite{Thrun:2005} are used to build maps of the robot's physical environment while simultaneously localizing the robot inside the map. In order to incorporate additional information into maps, such as semantics~\cite{Kostavelis:2015} or social information~\cite{Kruse:2013}, there are two different approaches: approaches based on implicit observations and explicit methods based on human-robot interaction. An example for the former category is the system by O'Callaghan et al. who adapt maps to human positional traces~\cite{O'Callaghan:2011}. Additional to human trajectory observations, Alempijevic et al. also exploit robot sensors to jointly learn a map for navigational purposes~\cite{Alempijevic:2013}. Ogawa et al. propose human motion maps to represent the distribution of human motion in a map~\cite{Ogawa:2014}. The location of passages is learned in the work of Papadakis et al. by observing human interactions in the spatial neighborhood~\cite{Papadakis:2016}. Since these approaches are based on observations, they are user-friendly because no explicit user interaction is necessary. However, they are not suited to flexibly define \textit{arbitrary} areas in an environment because they do not take the users' intentions into account. Therefore, we argue that such implicit methods based on observations are not appropriate for our problem. Thus, we focus on explicit approaches to incorporate additional information into maps.\par

An explicit approach is the work by Sakamoto et al. who propose a GUI-based interface to sketch the area for a vacuum cleaning task~\cite{Sakamoto:2016}. However, this approach requires several top-view cameras in the environment to establish a correspondence between the environment displayed on the GUI and the real environment. An alternative is the direct definition of virtual borders in a previously created occupancy grid map, that today's home robots already provide. The drawback of this method is the low accuracy since it is hard to establish correspondences between points in the map and the physical environment, especially if it is featureless. Commercial solutions for vacuum cleaning robots encompass magnetic stripes placed on the ground~\cite{Neato:2018} and virtual wall systems based on beacon devices~\cite{Chiu:2011}. These approaches are intrusive, power-consuming or inflexible. Sprute et al. purposely address these aspects by proposing a framework for teaching virtual borders based on robots' trajectories~\cite{Sprute:2017}. Although the framework is promising, the interaction method only supports polygons as virtual borders, employs visual markers to guide the mobile robot and was not evaluated with non-expert users. Thus, the method is not user-friendly and applicable in real life.\par

To address these aspects and because of the reasons mentioned in the introduction, we chose a laser pointer as interaction device. Laser pointers have been deployed in different scenarios as interaction devices, e.g. guiding a robot to a 3D location~\cite{Kemp:2008} or controlling a robot using stroke gestures~\cite{Ishii:2009}. Furthermore, Choi et al.~\cite{Choi:2008} investigate different interfaces for providing 3D locations, and results show that laser pointer interfaces are faster compared to touch screen interfaces. Conversely, Mikawa et al.~\cite{Mikawa:2010} use a laser pointer on a librarian robot to guide a human and indicate 3D positions.

\section{Notation \& Problem Definition}
The pose of a mobile robot operating in the 2D plane is defined as a triple $(x, y, \theta)$ with a current location $(x, y)$ and orientation~$\theta$. The robot's pose at a certain time $k$ is denoted as $x_{k}$, and the set of robot poses $\{x_{0}, x_{1}, ..., x_{k}\}$ describes the robot's pose history $X_{0:k}$ up to time $k$. A colored image obtained by the robot's camera is denoted as $\mathcal{I}_{RGB}(x, y) \in \mathbb{R}^3$ or $\mathcal{I}_{HSV}(x, y) \in \mathbb{R}^3$ depending on the color space. A depth image is defined analogously $\mathcal{I}_{Z}(x, y) \in \mathbb{R}$. A domain $\Omega(\mathcal{I}) \subset \mathbb{R}^2$ of an image or map $\mathcal{I}$ contains all possible (image) coordinates, thus $(x,y)^T \in~\Omega(\mathcal{I})$. We model the environment of a robot as a 2D occupancy grid map $M$. Each of the $m \times n$ cells represents a certain area in the corresponding environment and gives information about the occupancy probability. An occupancy value of a position $(x, y)$ of the map $M$ is denoted as $M(x, y)$.\par

In this work, we want to integrate virtual borders into a given map $M_{prior}$ of the environment resulting in a posterior map $M_{posterior}$. The prior map is a static or previously learned posterior map, while the posterior map contains the physical environment as well as the user-defined virtual borders from the interactive teaching process. This map can be used in future navigation tasks as basis for a global costmap.

\section{Virtual Border Teaching Using a Pointer}
\label{sec:teaching}
Due to the lack of flexible and user-friendly solutions to our challenge, we propose an interaction method using a laser pointer to teach virtual borders to mobile robots. It is based on a previously developed framework~\cite{Sprute:2017} that will be extended to provide a more flexible way to teach virtual borders. Thus, a user can interactively define \textit{arbitrary} workspaces for his or her mobile robots. Furthermore, the method only uses a robot's on-board sensors in addition to the interaction device. The overall goal is to change the robot's navigational behavior according to the users' needs. A full video of a teaching process can be found at: \url{https://youtu.be/lKsGp8xtyIc}.

\subsection{Interaction Method}
\label{sec:concept}
\begin{figure}
	\centering
	\includegraphics[width=0.44\textwidth]{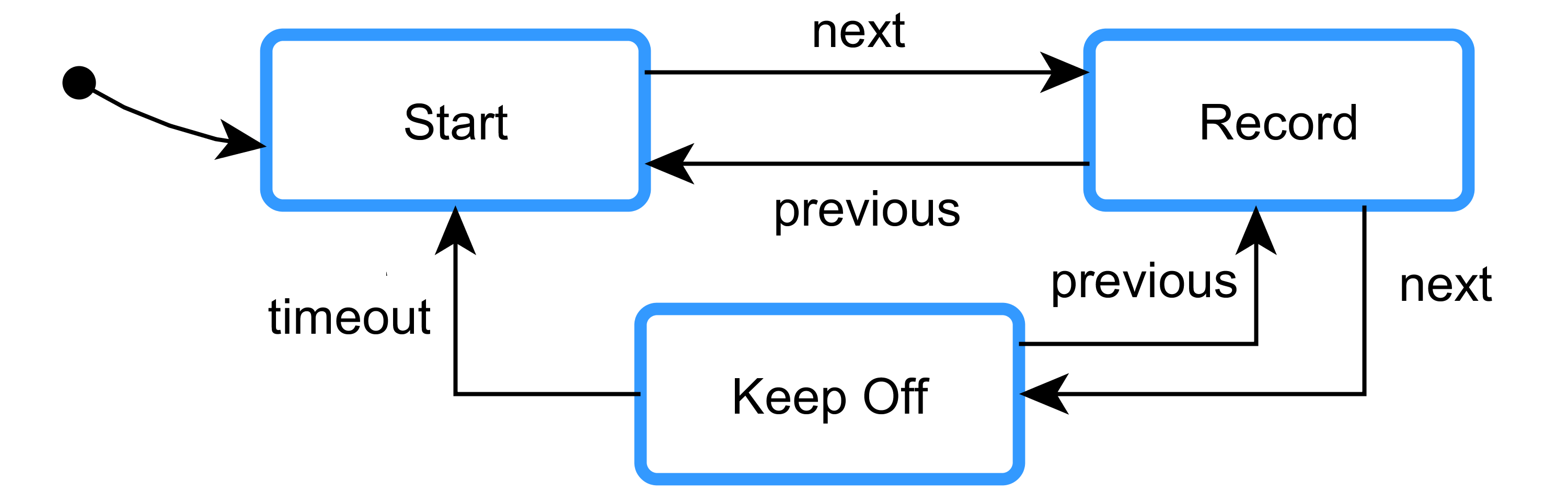}   
	\caption{State machine showing the different states and transitions of the interaction method.}
	\label{fig:statechart}
\end{figure}

The user employs a laser pointer to guide the robot along the desired virtual border while the robot simultaneously keeps track of its pose data $X_{a:b}$. The interaction method comprises three different states that we will refer to throughout the paper:
\begin{enumerate}
	\item \textit{Start} The user guides the robot to a start position for border learning by employing the laser pointer. The mobile robot follows the laser spot on the ground. 
	\item \textit{Record} The robot continues following the laser spot and simultaneously records its pose history $X_{a:b}$ from the time $a$ entering the state until leaving the state~$b$.	
	\item \textit{Keep Off} The mobile robot stops following the laser spot, and the robot's pose history $X_{a:b}$ is used to partition the prior map into two areas. We support polygons and curves separating an area (\textit{separating curves}). Subsequently, the user has the possibility to rotate the robot around its  vertical axis to indicate the keep off area which will not be intruded by the robot in future navigation tasks. Finally, the virtual border is integrated into the prior map.
\end{enumerate}
Fig. \ref{fig:statechart} gives an overview of the different states and transitions, that are triggered by user interactions using visual codes generated by the pointer or push buttons on the mobile robot. One may argue that it would be more intuitive to ``draw'' the virtual borders on the ground using the laser pointer instead of guiding the robot along the borders. Although this may be true, we purposely realized it this way because of two reasons. First, due to the limited field of view of the robot's camera, the robot needs to follow the laser spot to record the borders. Moreover, if the robot's trajectory defines a virtual border, it directly gives feedback to the user since the robot moves along the border. This would not be the case if the robot would only use the positions of the laser spot as border definition.

The process of incorporating virtual borders into a prior map is subdivided into the following tasks: (1)~laser point detection and following, (2)~user interaction to switch between states and (3)~the creation of the posterior map. We detail these tasks in the following subsections.

\subsection{Laser Point Detection \& Following}
\begin{figure*}
	\centering
	\includegraphics[width=\textwidth]{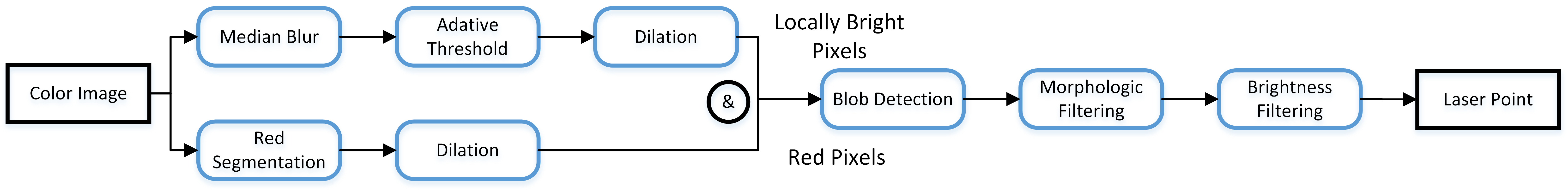}   
	\caption{Image processing pipeline for laser point detection. The input is a color image, and the result is the 2D position of the laser point in the image plane if present.}
	\label{fig:pipeline}
\end{figure*}

The first step is the detection of a laser point in an image. Therefore, a front-mounted RGB-D camera on the mobile robot acts as primary sensor and captures images of the scene. The user generates a red laser spot on the ground using a laser pointer. This laser spot has several properties that will be addressed in the detection process:
\begin{enumerate}
	\item The spot's main color is red.
	\label{enum:prop1}
	\item The spot is brighter than its surrounding environment.
	\label{enum:prop2}
	\item The spot has a size of approximately 5 mm $\times$ 5 mm depending on the material of the underground.
	\label{enum:prop3}
	\item The spot is approximately circular.
	\label{enum:prop4}
\end{enumerate}
We apply a multi-stage image processing approach to detect the laser point in the input image $\mathcal{I}_{RGB}$. The processing pipeline is tailored to the characteristics of a laser point and is shown in Fig. \ref{fig:pipeline}. 
The first steps of the image processing pipeline are processed in parallel to identify locally bright and red areas (see properties~\ref{enum:prop1} and ~\ref{enum:prop2}). The bit-wise conjunction of both processed images results in a mask that contains red and locally bright pixels. In order to extract laser point candidates $C \in \Omega(\mathcal{I}_{RGB})$, blob detection is performed on the combined image. Afterwards, blobs are discarded that do not match the morphologic characteristics of a laser point, i.e. the size (see property~\ref{enum:prop3}) and the circularity (see property~\ref{enum:prop4}) of the blob. Finally, the brightness of a blob center $(x_c, y_c)$ represented by the $V$-value of $\mathcal{I}_{HSV}(x_c, y_c)$ has to exceed a certain threshold to remain a laser point candidate. If more than one blob fulfills all the criteria, the brightest candidate point $\bm{l}$ is chosen: 
\begin{equation}
\bm{l} = \underset{(x_c, y_c) \in C}{\arg\max}\ V(\mathcal{I}_{HSV}(x_c, y_c))
\end{equation}
The $V(\cdot)$-operator extracts the $V$-value of $\mathcal{I}_{HSV}(x_c, y_c)$.\par

To follow the laser point detected in the input image, its 2D image coordinate $\bm{l} = (x, y)$ is transformed into 3D space using the inverse projection of the pinhole camera model: 
\begin{equation}
\bm{L} = \pi^{-1}(\bm{l}, Z) = \left( \dfrac{x - c_x}{f_x} Z, \dfrac{y - c_y}{f_y} Z, Z \right)^T
\end{equation}
$f_x$ and $f_y$ are the focal lengths in pixels, and $(c_x, c_y)$ is the principal point in image coordinates. These intrinsic camera parameters are obtained during a calibration process. The distance to the camera is denoted as $Z = \mathcal{I}_{Z}(\bm{l})$. After transforming the image coordinates of the laser point $\bm{l}$ into space $\bm{L}$, the robot can follow the laser point using visual servoing technique. The distance information is used to adjust the robot's velocity and to stop the robot if the distance falls below a certain threshold.

\subsection{User Interaction \& Feedback}
A user can switch between different states of the system by either using visual codes generated by the laser pointer or push buttons on the mobile robot platform. The former one is the more comfortable one, while the latter one is the robuster one. We chose this multimodal interaction since the interaction using visual codes can be error-prone under certain light conditions. In this case, a user can easily use the robot's on-board push buttons to ensure the functionality of the system. In the concrete implementation, two different visual codes or buttons are sufficient to realize the transition events \textit{next} and \textit{previous} according to the state machine in Fig. \ref{fig:statechart}. Apart from these user interactions, we also provide a simple feedback system that does not rely on additional hardware. The system signalizes its internal state using colored LEDs on the mobile robot, each color corresponding to one of the three system's states. Additional to a color change of the LED in case of a state change, the mobile robot employs a sound feedback (beep tones) to signalize the state change. Finally, our choice for a laser pointer as interaction device aims to provide users a direct visual feedback of their interaction.

\subsection{Map Creation}
Map creation is the task of integrating virtual borders into a given map. It depends on a given prior map $M_{prior}$, e.g. a static or a previously learned posterior map, the robot's pose history $X_{a:b}$ in the state \textit{Record} depending on the time interval $[a, b]$ and the last known pointer location $\bm{L}$ with respect to the map coordinate frame. First, we extract the robot's positions $\bm{p}_i \in \mathbb{R}^2$ from the robot's pose history $X_{a:b}$ because they are used to define the virtual borders. The result is a polygonal chain $\mathcal{P}$ consisting of $n$ points. We support two kinds of borders to flexibly define arbitrary virtual borders:
\begin{enumerate}
	\item Simple polygonal chains: If the Euclidean distance between the first $\bm{p}_1$ and the last $\bm{p}_n$ point of the polygonal chain $\mathcal{P}$ exceeds a certain threshold, the polygonal chain is considered as a simple polygonal chain.
	\item Closed polygonal chains: If the Euclidean distance between the first $\bm{p}_1 $ and the last $\bm{p}_n$ point of the polygonal chain $\mathcal{P}$ falls below the threshold, the polygonal chain is considered as a closed polygonal chain.
\end{enumerate}
To update the given map $M_{prior}$ with virtual borders, we use the polygonal chain $\mathcal{P}$ to partition the map into two areas:
\begin{equation}
A_c = \{\bm{c} \in \Omega(M_{prior}) \mid \bm{c}\ connected\ to\ \bm{L}^* \},
\end{equation}
which is the area that is directly connected to the cell corresponding to the last known laser spot position $\bm{L}^*$. The complementary area is denoted as $A_{nc}$. In case of a simple polygonal chain, the first $[\bm{p_1 p_2}]$ and the last $[\bm{p_{n-1} p_n}]$ line segments of the polygonal chain $\mathcal{P}$ are linearized to allow the partitioning of the map. Finally, the posterior map is constructed as follows:
\begin{equation}
M_{posterior}(x, y)=
\begin{cases}
	Occupied& if\ (x, y) \in A_c\\
	M_{prior}(x, y)& if\ (x, y) \in A_{nc}
\end{cases}
\end{equation}
Thus, the last position of the laser point defines the keep off area in the map, which can be used in future navigation tasks as basis for a 2D global costmap to change the robot's navigational behavior. It is possible to repeat this process several times to obtain a map with arbitrary virtual borders.

\section{Evaluation}
We evaluated the proposed interaction method with respect to three criteria: correctness, accuracy and teaching time. Due to practical reasons, we conducted a separated single-user and a multiple-user evaluation to consider different aspects of a teaching process, i.e. border lengths, variations and number of users. Each evaluation focused on different aspects to complement each other. A TurtleBot v2 equipped with a laser scanner and a front-mounted \mbox{RGB-D} camera served as evaluation platform. The whole system was implemented as a ROS package, and we performed the following experiments on occupancy grid maps with a resolution of 2.5~cm per pixel. A prior map was created before evaluation.

\subsection{Correctness}
\begin{figure}
	\centering
	\includegraphics[width=0.495\textwidth]{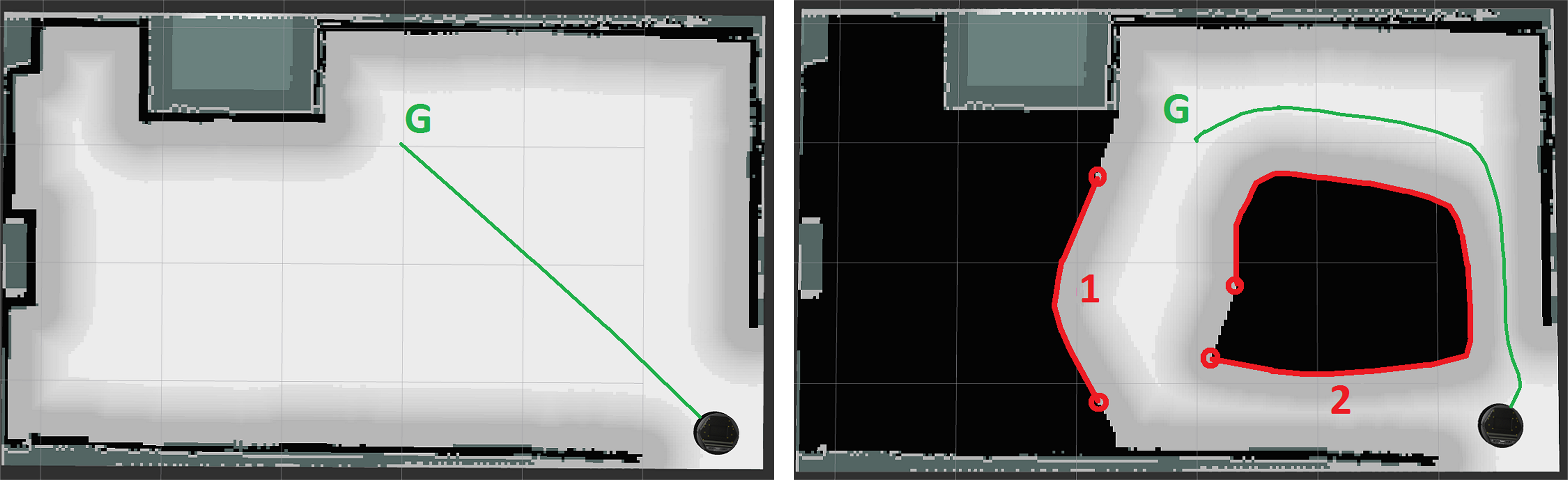}   
	\caption{Navigational costmap without / with virtual borders.}
	\label{fig:correctness}
\end{figure}

The first part of the evaluation is concerned with the correctness of the proposed system. It is the goal to give the user the possibility to define arbitrary virtual borders in an environment using a laser pointer. An example is shown in Fig.~\ref{fig:correctness}. First, a user defines a virtual border to exclude the left area of the map from the robot's workspace by drawing a separating curve with the laser pointer (see border~1). Afterwards, the user defines a virtual border around a carpet with the goal to declare the area of the carpet as $occupied$ (see border~2). Subsequently, the robot is given a navigation goal to move to a certain position in the map. The left costmap visualizes the global costmap and the path for the given navigation goal before teaching virtual borders. The robot moves from its current position directly to the navigation goal and crosses the carpet area. After teaching the virtual borders, the global costmap and path to the same navigation goal are shown in the right costmap. The robot now takes a different path to the goal and circumvents the carpet. This shows that the system can correctly incorporate different types of virtual borders that are respected in subsequent navigation tasks successfully.

\subsection{Single-User Evaluation}
This evaluation aims to analyze the effect of the border length and variations in the teaching process on the accuracy and teaching time. All experiments were performed by a non-expert user (male, 26 years) as defined in the introduction. The user had any time as needed to get familiar with the interaction method before performing the evaluation (approx. 5 minutes). 

\subsubsection{Accuracy}
\begin{figure}
	\centering
	\includegraphics[width=0.495\textwidth]{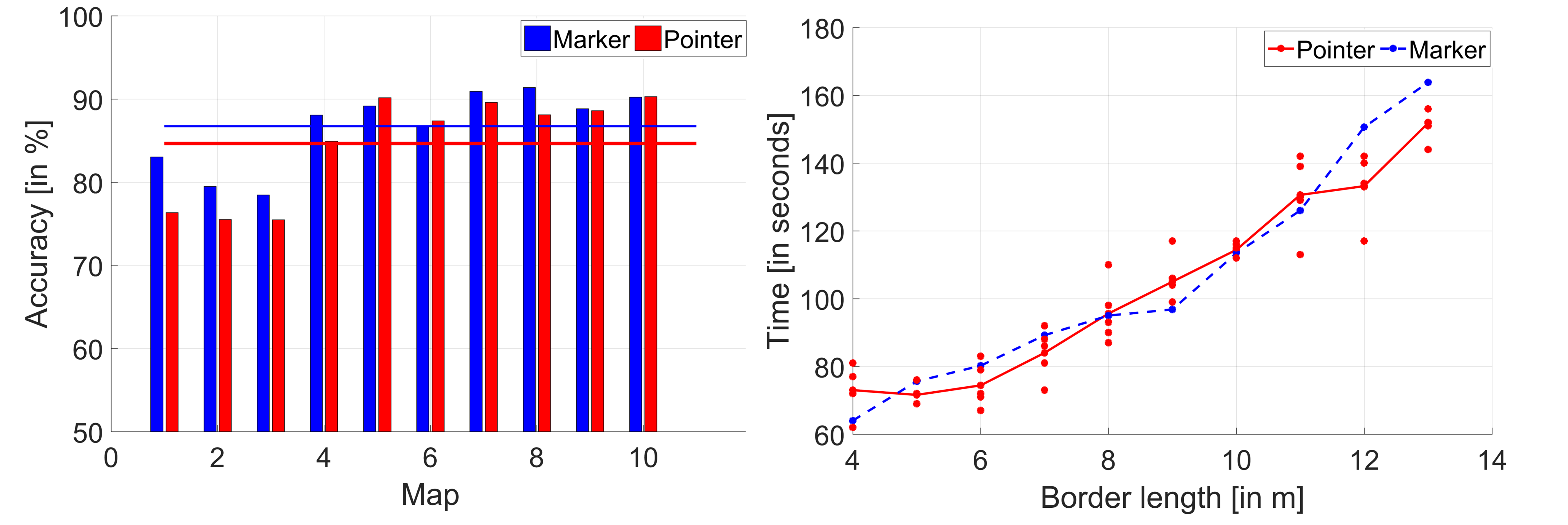}   
	\caption{Results of the accuracy and teaching time.}
	\label{fig:accTimeEval}
\end{figure}

In order to evaluate the accuracy of the interaction method, we compared resulting maps from several teaching processes with previously created maps containing virtual borders. For this purpose, we used a dataset of ten different maps of the 6.1~m $\times$ 3.5~m lab environment with different virtual borders integrated. The dataset contains virtual borders with a length of 4~m to 13~m and different complexities. These virtual borders were manually integrated into the existing map of the lab environment. Additionally, the virtual borders were marked in the physical environment to allow the user the teaching of the ground truth maps. Each map was defined five times by a user resulting in 50 runs in total. Five runs were performed for each map to introduce some variation in the teaching process, e.g. starting the $Record$ phase from different positions. Additionally, we compared the results using the laser pointer as interaction device with a previously developed marker-based approach~\cite{Sprute:2017}. It uses visual markers, each encoded with a different ID, to guide the robot and to switch between the states of the state machine. We chose this interaction method as baseline because it is the most flexible one from the literature and it is based on the same robot guidance framework. We applied the Jaccard index 
\begin{equation}
J(GT, UD) = \dfrac{|GT \cap UD|}{|GT \cup UD|} \in [0, 1]
\end{equation}
to assess the accuracy of the learned virtual borders. $GT$ is the set of cells in the map that belong to the ground truth virtual areas, whereas $UD$ is the set of cells in the map that belong to the user-defined virtual areas. The number of overlapping cells in $GT$ and $UD$ is denoted as $|GT \cap UD|$, and $|GT \cup UD|$ is the number of cells contained in the union set. As shown in the left of Fig.~\ref{fig:accTimeEval}, both approaches have high accuracy values (84.6\% average for pointer approach and 86.6\% average for marker approach) and do not significantly differ from each other. Small errors occur due to the interaction between human and robot and small localization errors of the robot. It is also apparent that the accuracies of maps 1-3 are less than the accuracies of maps 4-10. The former ones are maps with virtual borders ranging from 4~m to 6~m which makes it hard to guide the robot on such a small area. In contrast to that, the latter ones contain borders in the range of 7~m to 13~m allowing a more precise guidance  of the robot. Note that for reasons of comparability, we adapted the accuracy values reported by the authors in~\cite{Sprute:2017} according to our similarity index. They used a similar index but incorporated the physical environment into the calculation which makes it less comparable with other approaches performed in different physical environments.

\subsubsection{Teaching Time}
While conducting the accuracy runs, the time needed to teach the virtual borders was measured. Since the teaching time is affected by the length of the learned virtual border, we evaluated the teaching time depending on the border length. The teaching time is the time when the robot records its poses, i.e. the robot is in state $Record$. The results compared to the marker approach are shown in the right of Fig.~\ref{fig:accTimeEval}. It reveals a linear relationship between the teaching time and the border length. This is due to the nature of the framework: the robot follows the laser point and the longer the length of the border is, the more time it takes to drive along the border. Compared to the marker-based approach, there is no significant difference between both interaction devices. Additionally, the small variances of the data points indicate robustness towards the variations introduced in the teaching process.

\subsection{Multiple-User Evaluation}
Additional to the results for a single non-expert, we also evaluated multiple users' performances concerning the accuracy and teaching time in a typical use case of our interaction method. We placed a common carpet (2.00~m $\times$ 1.25~m) as shown in Fig.~\ref{fig:motivation} on the ground and created a ground truth map manually for this scenario. Afterwards, we instructed 15 non-experts (9 male, 6 female) to exclude the carpet area from the robot's workspace. They rated their robotic skills on an 11-point Likert scale with a mean of $M=1.33$ (1~=~low skill, 11~=~high skill). Each participant performed the experiment with the marker and laser pointer approach in a random order. The mean age of the participants was 35 years ranging from 16 to 56 years. They were all participants who were not involved in the design or implementation of the interaction method. An experimenter measured the teaching time of each participant and saved the resulting posterior maps after each run. These were compared to the ground truth map by calculating the Jaccard index. This evaluation aims to assess the accuracy and teaching time based on multiple non-experts' performances.\par

The results compared to the corresponding results of the single-user evaluation are shown in Table~\ref{tab:multiUser}. In case of the accuracy, there is a difference of approximately 20\% between the single-user and multiple-user evaluation, and our proposed method features a slightly better accuracy compared to the marker approach. In case of the teaching time, it is similar: the multiple-user results are worse than the single-user results, and the teaching time of our proposed method is better. Since the single user performed all runs on different maps and with variations, he got some experience in handling the interaction devices. In contrast to this, each of the 15 participants in the multiple-user evaluation defined virtual borders for the first time. Thus, experience in handling the interaction device can increase the accuracy and reduce the teaching time.

\begin{table}[htbp]
  \centering
  \caption{Quantitative results of the multiple-user compared to the single-user evaluation (mean $\pm$ standard deviation).}
    \begin{tabular}{l|rr|rr}
          & \multicolumn{2}{c}{Accuracy [in \%]} & \multicolumn{2}{c}{Time [in seconds]} \\
          & \multicolumn{1}{l}{Multiple user} & \multicolumn{1}{l}{Single user} & \multicolumn{1}{l}{Multiple user} & \multicolumn{1}{l}{Single user} \\
          \toprule
    Marker & 65.2 ($\pm$ 5.2) & 86.6 ($\pm$ 2.8) & 129 ($\pm$ 23)  & 85 ($\pm$ 5) \\
    Pointer & 66.4 ($\pm$ 8.3) & 84.6 ($\pm$ 3.5) & 112 ($\pm$ 24)  & 79 ($\pm$ 7)\\
    \end{tabular}%
  \label{tab:multiUser}%
\end{table}%

\subsection{Discussion}
Our experimental results showed the correctness of the interaction method, i.e. the user-defined virtual borders, polygon and separating curve, were correctly incorporated into the prior map of the environment and the mobile robot changed its navigational behavior. The results of the single-user evaluation revealed a linear relationship of the teaching time with respect to the length of the virtual border, and the accuracy is independent of the border length. However, the accuracy decreases for short virtual borders (4-6~m) because of the difficulties to guide a mobile robot on small areas. The results of the multiple-user evaluation showed a drop in accuracy (pointer: -18.2\%) and an increase of the teaching time (pointer: +33~s) compared to the single-user evaluation. This is caused by a learning effect. Comparing the results of the multiple-user evaluation, our laser pointer approach achieves a similar accuracy and a reduction of the teaching time (-17~s) compared to the marker approach. We observed that this is caused by the interaction device: participants had problems when guiding the robot with visual markers. In case of the laser pointer, all participants could easily employ the laser pointer and could guide the robot resulting in a reduced teaching time. With the choice of a laser pointer as human-robot interaction device, we purposely addressed this lack of the baseline's usability. 


\section{Conclusions \& Future Work}
We proposed an interaction method to flexibly teach virtual borders to mobile robots using a laser pointer. To allow flexible teaching of arbitrary virtual borders, we extended a previously developed teaching framework by considering different border types, i.e. polygons and separating curves. It allows non-expert users to effectively restrict the workspace of a mobile robot to certain areas and change the robot's navigational behavior. This is especially interesting for a human-aware navigation in human-centered environments. An experimental evaluation with non-expert users showed that our interaction method is applicable by non-expert users, features a high accuracy and linear teaching time with respect to the border length. Compared to the marker-based approach, our interaction method is more flexible and features a reduced teaching time and a more natural user interaction. Currently, feedback to the user is provided through sound, colored LEDs on the robot, the laser pointer's position and through the position of the mobile robot indicating the virtual border points. A limitation of this work is the missing capability to visualize the user-defined virtual borders. Therefore, future work should focus on the realization of an enhanced feedback system, e.g. using projectors mounted on the robot to visualize virtual borders. Finally, the interaction method needs to be extended to allow users to remove virtual borders if they are no longer needed.

\section*{Acknowledgment}
This work is financially supported by the German Federal Ministry of Education and Research (grant: 03FH006PX5).



\bibliographystyle{IEEEtran}
\bibliography{bibo}

\end{document}